\documentclass[runningheads]{llncs}
\usepackage[T1]{fontenc}
\usepackage{graphicx}

\usepackage{color}
\usepackage{float}
\usepackage{booktabs}
\usepackage{subcaption}
\usepackage{amsmath} 
\definecolor{myred}{rgb}{.8,.0,.0}

\usepackage{url}
\usepackage{hyperref}

\ifdefined\DOUBLEBLIND
\newcommand{\repourl}{\url{https://anonymous.4open.science/r/clip_cxr_fairness-BCCC}}
\else
\newcommand{\repourl}{\url{https://github.com/TheoSourget/clip_cxr_fairness}}
\fi

\begin{document}
\title{Fairness and Robustness of CLIP-Based Models for Chest X-rays}

\ifdefined\DOUBLEBLIND
    \author{***}
    \authorrunning{***}
    \institute{***}
\else
    \author{Théo Sourget\inst{1} \and
    David Restrepo\inst{1} \and
    Céline Hudelot\inst{1} \and
    Enzo Ferrante\inst{2} \and
    Stergios Christodoulidis\inst{1} \and
    Maria Vakalopoulou\inst{1}}
    \authorrunning{T. Sourget et al.}
    \institute{
    MICS, CentraleSupélec - Université Paris-Saclay, France\\
    \email{\{theo.sourget, david.restrepo, celine.hudelot, stergios.christodoulidis, maria.vakalopoulou\}@centralesupelec.fr}\and
    CONICET, Universidad de Buenos Aires, Argentina\\
    \email{eferrante@sinc.unl.edu.ar}
    }
\fi

\maketitle              
\begin{abstract}
Motivated by the strong performance of CLIP-based models in natural image-text domains, recent efforts have adapted these architectures to medical tasks, particularly in radiology, where large paired datasets of images and reports, such as chest X-rays, are available. While these models have shown encouraging results in terms of accuracy and discriminative performance, their fairness and robustness in the different clinical tasks remain largely underexplored. In this study, we extensively evaluate six widely used CLIP-based models on chest X-ray classification using three publicly available datasets: MIMIC-CXR, NIH-CXR14, and NEATX. We assess the models fairness across six conditions and patient subgroups based on age, sex, and race. Additionally, we assess the robustness to shortcut learning by evaluating performance on pneumothorax cases with and without chest drains. Our results indicate performance gaps between patients of different ages, but more equitable results for the other attributes. Moreover, all models exhibit lower performance on images without chest drains, suggesting reliance on spurious correlations. We further complement the performance analysis with a study of the embeddings generated by the models. While the sensitive attributes could be classified from the embeddings, we do not see such patterns using PCA, showing the limitations of these visualisation techniques when assessing models.
Our code is available at \repourl

\end{abstract}

\section{Introduction}
\label{sec:intro}
Deep learning models that have been trained on large-scale chest X-ray datasets have achieved performances reaching expert-levels in disease classification of X-ray images~\cite{hosny2018artificial,shen2019artificial}. However, despite such models obtaining strong benchmark performance, different studies show that they often exhibit performance disparities across patient subgroups, revealing concerning biases. For example, studies like \cite{glocker2023algorithmic,larrazabal2020gender} show how the performance of convolutional neural networks (CNN) can vary based on demographic attributes such as age, sex, or race, particularly for X-rays.
Beyond performance differences, there is growing evidence that deep learning models encode sensitive demographic information in their internal representations. Previous studies like ~\cite{bahre2024fairness,restrepo2024df} are able to predict sensitive attributes from the embedding generated by pretrained models. Similarly, Gichoya et al.~\cite{gichoya2022ai} demonstrate that deep learning models can classify patient race, even when the input images are heavily corrupted, raising serious concerns about the implicit encoding of sensitive information and its fairness implications.

Additionally, other studies show the impact of artefacts, also called shortcuts, in the classification. Jiménez-Sánchez et al.~\cite{jimenez2023detecting} and Oakden et al.~\cite{oakden2020hidden} show that models for pneumothorax classification have lower performances on images without chest drains, a common treatment for this disease. Moreover, Sourget et al.~\cite{sourget2025mask} demonstrate the ability of the models to obtain good performances in chest X-rays classification while masking out the lungs in the image, showing how these models can rely on non-relevant features.

More recently, advances in multimodal and foundation models have led to the development of contrastively trained architectures that jointly leverage chest X-rays and radiology reports~\cite{bannur2023learning,boecking2022making,codella2024medimageinsight,tiu2022expert,wang2022medclip,you2023cxr}.  
While these vision-language models (VLMs) have demonstrated promising results, recent studies have raised concerns regarding the fairness of VLMs. Luo et al.~\cite{luo2024fairclip} assess the fairness of the original CLIP and BLIP2 models on glaucoma classification pretrained with both natural domain and medical data, showing differences across subgroups especially on the natural domain models.
Yang et al.~\cite{yang2025demographic} evaluate the fairness of the CheXzero model~\cite{tiu2022expert} for chest X-ray classification, showing the gap in performances between different subgroups. Finally, Fay et al.~\cite{fay2025beyond} compare the performances of multiple zero-shot and training-based strategies for the MedImageInsight model~\cite{codella2024medimageinsight} on pneumonia classification and include an assessment of their fairness, showing that zero-shot techniques present less bias compared to linear probing but still higher than with LoRA or \textit{k}-NN.

In this work, we extend these studies by evaluating a large set of CLIP-based models pretrained on X-ray data, providing complementary analysis of the embedding representations, and assessing the robustness of the models to shortcut learning. Our empirical study, aiming at improving the understanding of the biases of CLIP-based models for chest X-rays classification: 
\textbf{1)} evaluates the performance of six widely used CLIP-based vision-language models on the multilabel classification of chest X-rays; \textbf{2)} assesses the fairness of the architectures on multiple subgroups of patients; \textbf{3)} studies the potential encoding of sensitive attributes in the embedding of these models with visualisation and classification techniques; \textbf{4)} compares their robustness regarding shortcuts on pneumothorax classification with and without chest drains.


\section{Fairness and Robustness of CLIP-based Models}
\subsection{Evaluation protocol on zero-shot classification}
We evaluate the performance of the models on different subgroups in a zero-shot classification setting. Inspired by the setups of \cite{fay2025beyond,tiu2022expert}, we compute for each label the similarities between the embeddings of an image and two templates "Chest \{CLASS\}" and "Chest No Findings". We then apply a softmax function between the two similarities to obtain the probability of the disease.
We evaluate the models across individual diseases and demographic subgroups (sex, race, and age) to assess both overall discriminative performance and subgroup fairness, quantified by performance disparities. We also assess whether the models rely on non-clinically relevant image features. To this end, we evaluate their performance for pneumothorax classification on two groups: one in which all patients with pneumothorax have chest drains, and the other in which they never have one.
Finally, we compute calibration curves for this task using the softmax values from the zero-shot classification to examine the reliability of the predicted probabilities.

We use the area under the receiver operating characteristic (AUC) and the adjusted area under the precision-recall curve (AUPRC$_{adj}$), which is usually adopted to evaluate models in a highly imbalanced scenario ~\cite{mosquera2024class}. The AUPRC$_{adj}$ is defined as $1-\frac{log(\text{AUPRC})}{log(\text{AUPRC}_{rng})}$ with $\text{AUPRC}_{rng}$ being the ratio between the number of positive samples for a class and the total number of samples.

\subsection{Encoding of sensitive attributes in the embedding space}
To further understand how these models work and what they learn in this multimodal contrastive setting, we generate and visualise the obtained image and text embeddings. 
For textual embeddings, as the text encoders have a limited input size, we only use the "FINDINGS" section from radiology reports, likely to contain the most relevant information. 
To assess the encoding of sensitive attributes, we use PCA to project the embeddings in two dimensions, revealing potential patterns with respect to patient sex, race, and age. We also train a model to classify the different sensitive attributes from the image embeddings using simple models like a linear probe (LP), a \textit{k}-nearest neighbours (\textit{k}-NN) classifier and a single-hidden-layer multi-layer perceptron (MLP). We split the original test sets in train, validation, and test subsets, ensuring that all images from a given patient are assigned to the same split to prevent data leakage. We used the validation set to tune the models' hyperparameters: the learning rate in the linear probe, the number of nearest neighbours, and the number of neurons of the MLP hidden layer.

Finally, following the analysis by Schrodi et al.~\cite{schrodi2025two} on the modality gap — which shows that differences between image and text embedding centroids are concentrated in a few dimensions — we conduct a similar analysis across patient subgroups. Specifically, we compute the centroid of image embeddings for each subgroup and measure the per-dimension differences between pairs of subgroup centroids.

\subsection{Data}
The MIMIC-CXR\footnote{Downloaded from \url{https://physionet.org/content/mimic-cxr-jpg/2.1.0/} and complemented with MIMIC-IV: \url{https://physionet.org/content/mimiciv/3.1/}}~\cite{johnson2019mimic,johnson2019mimicphysionet} dataset contains chest X-rays and radiology reports from 227,835 radiographic studies. Following standard practices in the training and evaluation of foundation models, we only use the original test split containing 30,359 images to avoid potential data leakage. Since some of our analyses need the "FINDING" section of the report, we only kept the 8950 samples for which this section is available in the test set. We use a subset of the classes available in the dataset: atelectasis, cardiomegaly, consolidation, pleural effusion, pneumonia, and pneumothorax.

The NIH-CXR14 dataset\footnote{Version 3 downloaded from \url{https://www.kaggle.com/datasets/nih-chest-xrays/data}}~\cite{Wang2017chestxray8} contains 112,120 X-ray images from 30,805 unique patients. We only use the 25,596 images of the test set. While the dataset contains annotations of 14 different conditions, here we focus on pneumothorax for our shortcut learning analysis.
The NEATX dataset\footnote{Version 1.0 downloaded from \url{https://zenodo.org/records/14944064}}~\cite{cheplygina2025NEATX,damgaard2023augmenting} contains annotations of chest drains in X-rays from the NIH-CXR14 and PadChest datasets. We use the annotations for the NIH-CXR14 dataset to assess the robustness of models to chest drains in pneumothorax classification. As the dataset only contains annotations of chest drains in positive samples of pneumothorax, using the hyperparameters described in the dataset paper~\cite{damgaard2023augmenting}, we train a DenseNet model for the detection of chest drains and automatically generate the labels for non-pneumothorax samples.

\subsection{Models}
We conduct our experiments with six CLIP-based architectures for which pretrained weights were available: MedCLIP~\cite{wang2022medclip}, Biovil~\cite{boecking2022making} and Biovil-t~\cite{bannur2023learning}, MedImageInsight~\cite{codella2024medimageinsight}, CheXzero~\cite{tiu2022expert}, and CXR-CLIP~\cite{you2023cxr}. All of these models were trained on datasets containing chest X-rays either exclusively or with other medical image modalities. We selected these models due to their recent release and their wide usage as baseline in previous works.

\section{Results}
\subsection{Good overall performances with subgroup-specific variability}

Table~\ref{tab:zeroshot_perf} shows the AUC and AUPRC$_{adj}$ of the different models on the MIMIC-CXR test set.  
One can see that aside from CXR-CLIP, the models obtain better than random values, especially for MedCLIP, MedImageInsight, and CheXzero, confirming their application in zero-shot settings.

\begin{figure}[ht]
    \centering
    \includegraphics[width=0.97\linewidth]{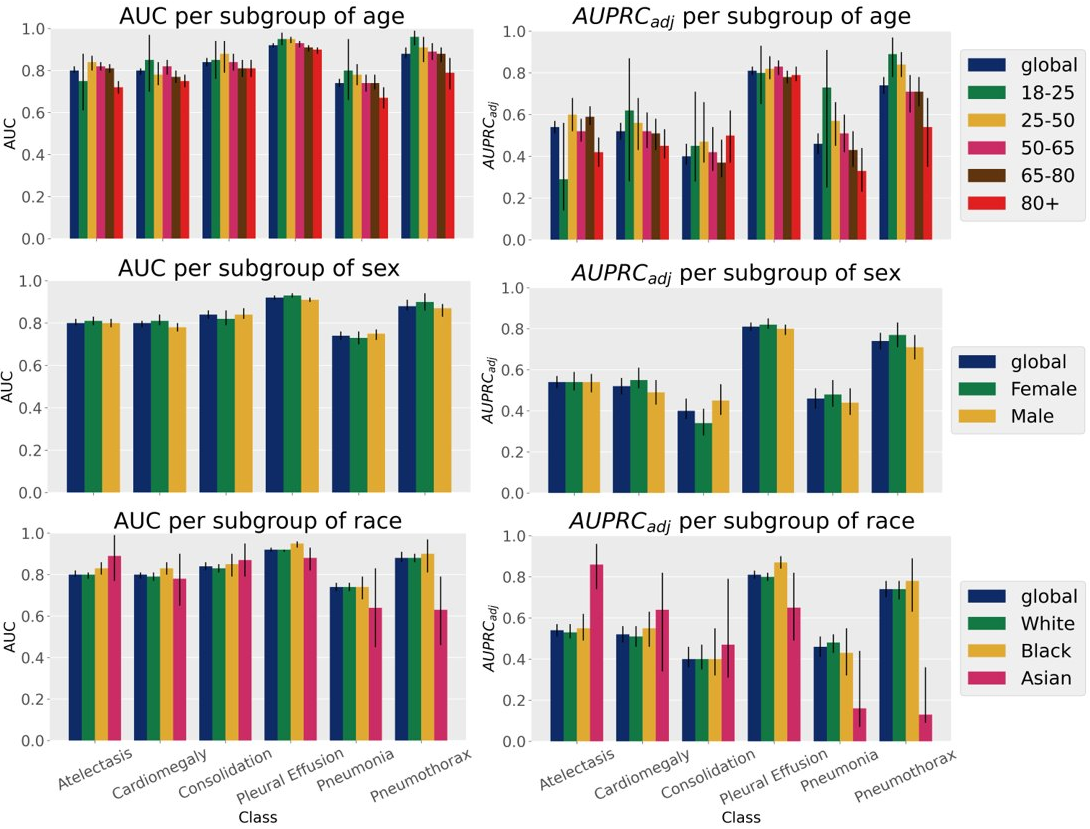}
    \caption{Example of AUC and AUPRC$_{adj}$ for the MedCLIP model and different subgroups with 95\% confidence interval using the bootstrap method with 1000 resamples}
    \label{fig:fairness_auc_auprc}
\end{figure}

For further evaluation, we generate for each model a barplot of the AUC and AUPRC$_{adj}$ per subgroup to observe potential gaps, see the results for the MedCLIP model in Fig.~\ref{fig:fairness_auc_auprc} with 95\% confidence intervals computed using the bootstrap method. While the results vary across the models and subgroups, we can still see a similar pattern with gaps across patient ages. The gaps seem, however, smaller for patient sex and race with the exception of Asian patients for which we can often see either a high improvement or decrease. However, this may be explained by the limited amount of positive samples per class for Asian patients leading to more extreme values and confidence intervals. Note that the same observation can be made for the 18-25 year old subgroup. It highlights the need for a more diverse test dataset to better estimate the true performance of the models on these subgroups.

\begin{table}[ht]
    \centering
        \resizebox{\textwidth}{!}{%
	\begin{tabular}{c|cc|cc|cc|cc|cc|cc|cc|}
	& \multicolumn{2}{c|}{\textbf{Atelectasis}} & \multicolumn{2}{c|}{\textbf{Cardiomegaly}} & \multicolumn{2}{c|}{\textbf{Consolidation}} & \multicolumn{2}{c|}{\textbf{Effusion}} & \multicolumn{2}{c|}{\textbf{Pneumonia}} & \multicolumn{2}{c|}{\textbf{Pneumothorax}} & \multicolumn{2}{c|}{\textbf{Mean}} \\
	& AUC & $\text{AUPRC}_{adj}$ & AUC & $\text{AUPRC}_{adj}$ & AUC & $\text{AUPRC}_{adj}$ & AUC & $\text{AUPRC}_{adj}$ & AUC & $\text{AUPRC}_{adj}$ & AUC & $\text{AUPRC}_{adj}$ & AUC & $\text{AUPRC}_{adj}$ \\ \hline
	MedCLIP & \textbf{0.8} & \textbf{0.54} & 0.8 & 0.52 & \textbf{0.84} & \textbf{0.4} & \textbf{0.92} & \textbf{0.81} & \textbf{0.74} & \textbf{0.46} & \textbf{0.88} & \textbf{0.74} & \textbf{0.83} & \textbf{0.58} \\
	& [0.79,0.82] & [0.51,0.57]& [0.78,0.81] & [0.48,0.56]& [0.82,0.86] & [0.36,0.46]& [0.91,0.93] & [0.79,0.83]& [0.72,0.76] & [0.41,0.51]& [0.86,0.91] & [0.7,0.78]& $\pm$ 0.06 & $\pm$ 0.16 \\
	Biovil & 0.68 & 0.2 & 0.76 & 0.41 & 0.42 & -0.06 & 0.69 & 0.38 & 0.49 & -0.0 & 0.72 & 0.21 & 0.63 & 0.19 \\
	& [0.66,0.69] & [0.18,0.23]& [0.74,0.77] & [0.36,0.45]& [0.39,0.46] & [-0.08,-0.03]& [0.67,0.7] & [0.35,0.42]& [0.47,0.52] & [-0.03,0.03]& [0.69,0.74] & [0.18,0.25]& $\pm$ 0.14 & $\pm$ 0.19 \\
	Biovil-t & 0.64 & 0.15 & 0.74 & 0.3 & 0.59 & 0.06 & 0.79 & 0.49 & 0.61 & 0.14 & 0.66 & 0.17 & 0.67 & 0.22 \\
	& [0.63,0.66] & [0.13,0.18]& [0.73,0.76] & [0.27,0.34]& [0.55,0.62] & [0.03,0.11]& [0.78,0.8] & [0.46,0.52]& [0.58,0.63] & [0.11,0.19]& [0.63,0.69] & [0.13,0.21]& $\pm$ 0.08 & $\pm$ 0.15 \\
	MedImageInsight & 0.74 & 0.36 & \textbf{0.85} & \textbf{0.53} & 0.83 & \textbf{0.4} & 0.88 & 0.7 & 0.69 & 0.33 & \textbf{0.88} & 0.63 & 0.81 & 0.49 \\
	& [0.73,0.75] & [0.33,0.39]& [0.83,0.86] & [0.49,0.57]& [0.8,0.85] & [0.35,0.46]& [0.87,0.89] & [0.67,0.72]& [0.67,0.72] & [0.29,0.38]& [0.86,0.9] & [0.58,0.68]& $\pm$ 0.08 & $\pm$ 0.15 \\
	CheXzero & 0.67 & 0.21 & \textbf{0.85} & 0.6 & 0.8 & 0.34 & 0.88 & 0.7 & 0.68 & 0.28 & 0.8 & 0.36 & 0.78 & 0.42 \\
	& [0.65,0.68] & [0.19,0.25]& [0.83,0.86] & [0.57,0.64]& [0.78,0.83] & [0.29,0.4]& [0.87,0.89] & [0.68,0.72]& [0.66,0.7] & [0.24,0.33]& [0.78,0.82] & [0.31,0.43]& $\pm$ 0.09 & $\pm$ 0.19 \\
	CXR-CLIP & 0.61 & 0.18 & 0.55 & 0.06 & 0.48 & -0.01 & 0.67 & 0.35 & 0.48 & -0.02 & 0.45 & -0.01 & 0.54 & 0.09 \\
	& [0.59,0.63] & [0.15,0.22]& [0.53,0.58] & [0.03,0.09]& [0.46,0.49] & [0.0,-0.0]& [0.66,0.69] & [0.31,0.38]& [0.46,0.5] & [0.0,0.02]& [0.41,0.48] & [0.0,0.09]& $\pm$ 0.09 & $\pm$ 0.15 \\
	\hline
	\end{tabular}
    }
	\caption{AUC and AUPRC$_{adj}$ of zeroshot classification. Negative AUPRC$_{adj}$ values denote results below the random classifier. Values in [] are the 95\% confidence intervals computed with the bootstrap method. $\pm$ in the Mean column are the standard deviations.}
	\label{tab:zeroshot_perf}
\end{table}

\subsection{Sensitive attributes are encoded in embeddings despite unclear visual separation}
\begin{figure}[ht]
    \centering
    \begin{subfigure}{.22\columnwidth}
        \centering
         \includegraphics[width=\linewidth]{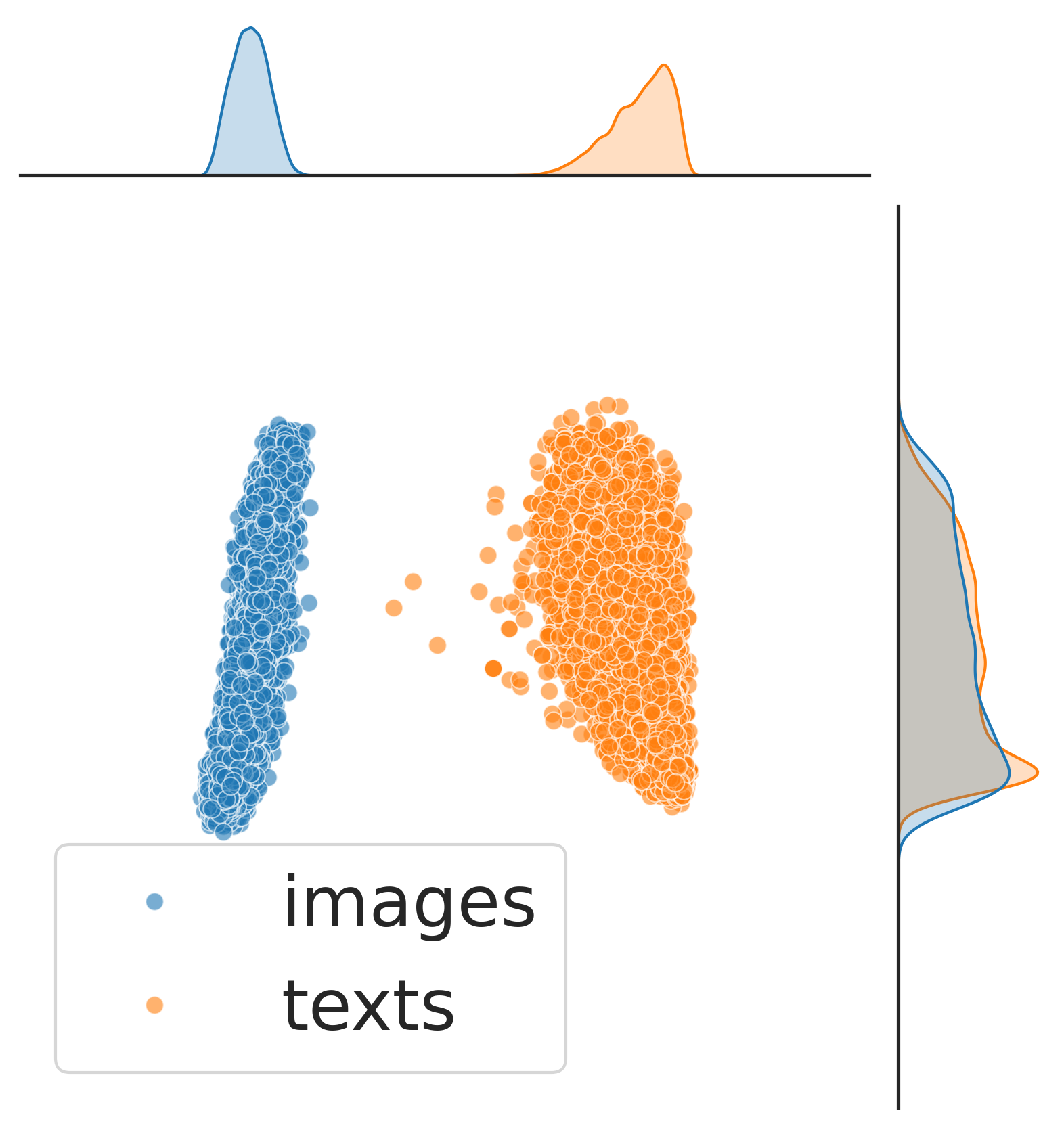}
      \caption{Modality}
      \label{subfig:modality_gap_pca}
    \end{subfigure}
    \begin{subfigure}{.22\columnwidth}
        \centering
        \includegraphics[width=\linewidth]{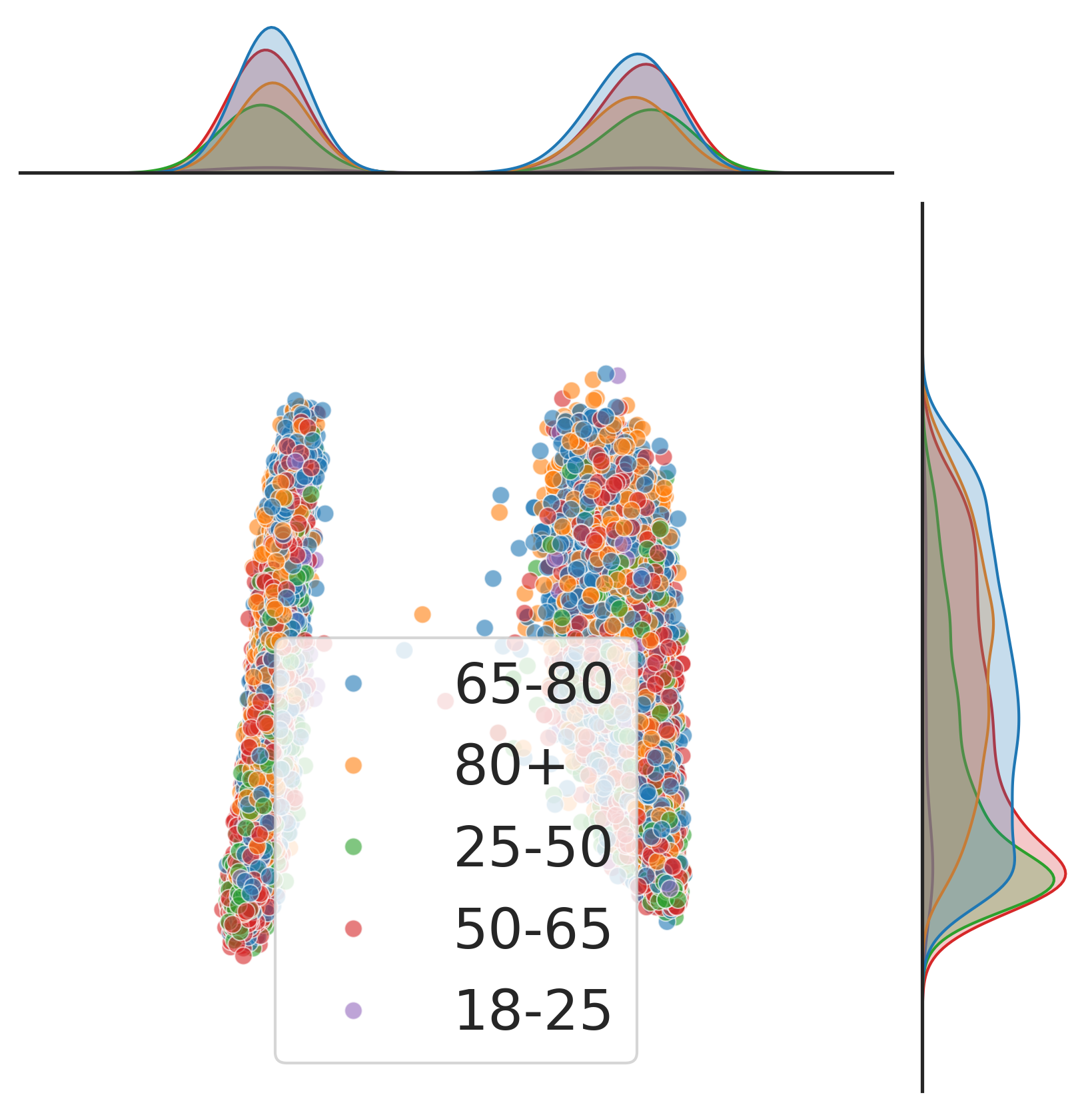}
        \caption{Patient age}
        \label{subfig:pca_age}
    \end{subfigure}    
    \begin{subfigure}{.22\columnwidth}
        \centering
        \includegraphics[width=\linewidth]{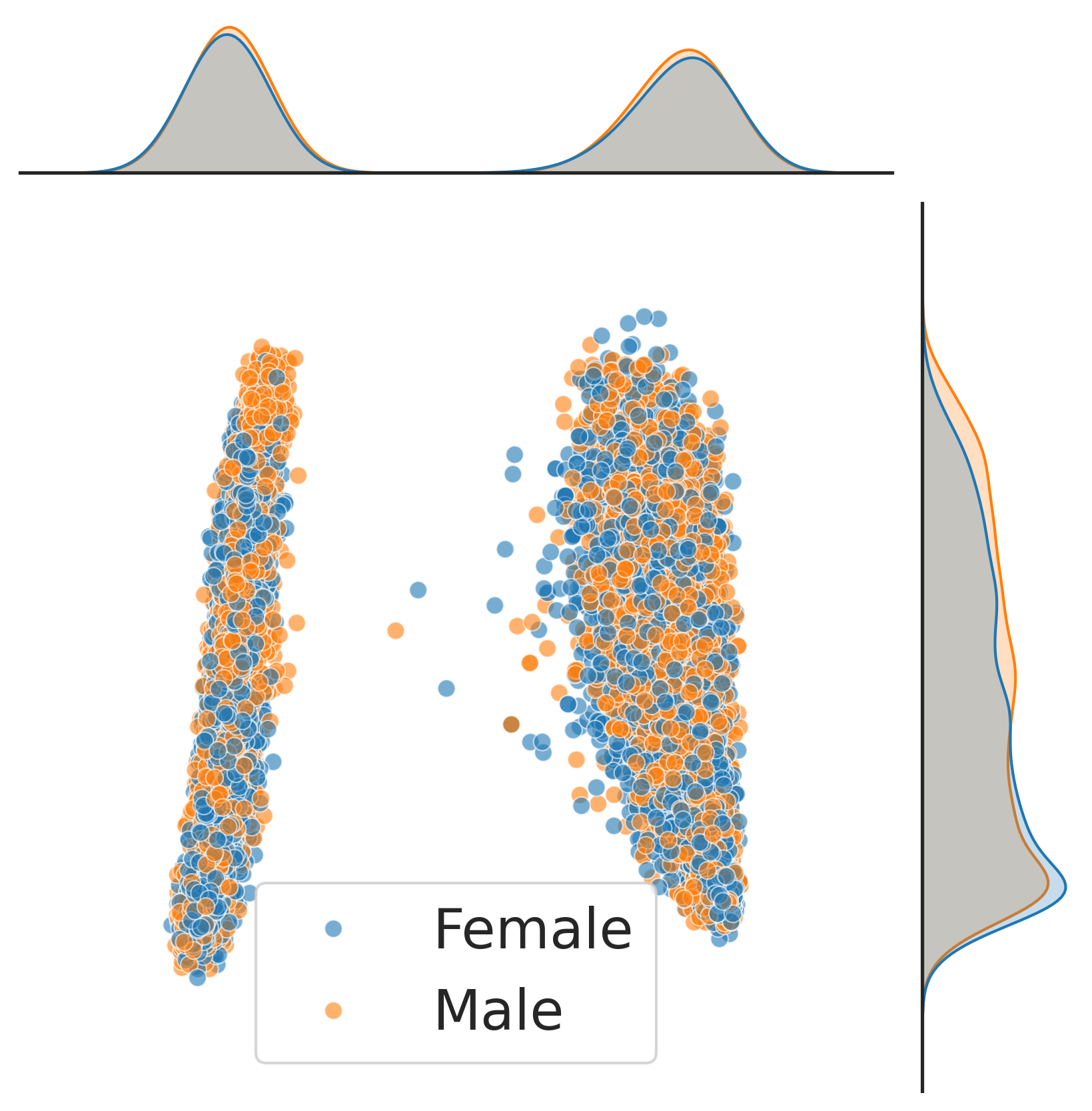}
        \caption{Patient sex}
        \label{subfig:pca_sex}
    \end{subfigure}
    \begin{subfigure}{.22\columnwidth}
        \centering
        \includegraphics[width=\linewidth]{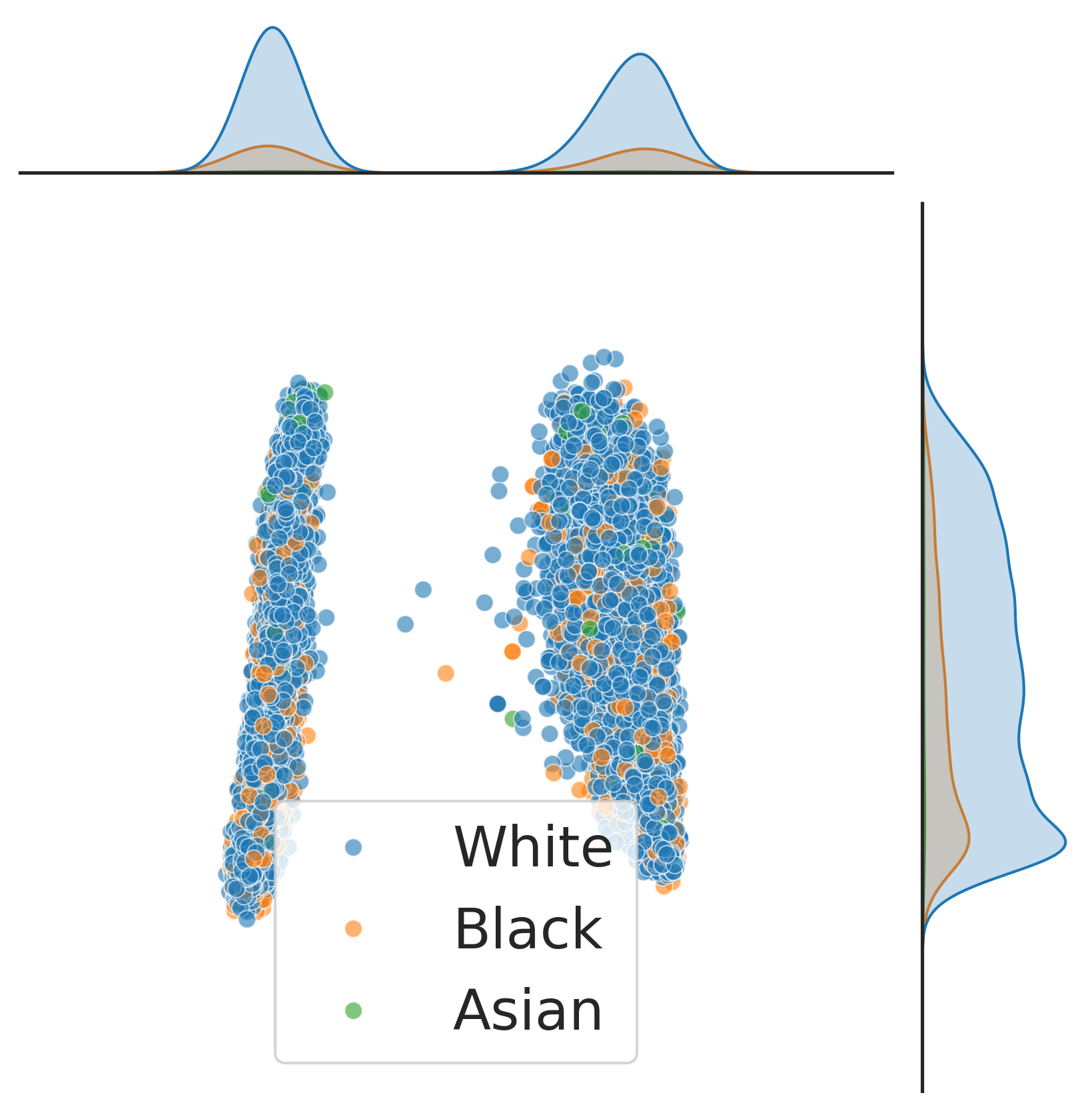}
        \caption{Patient race}
        \label{subfig:pca_race}
    \end{subfigure}
    \caption{PCA of MedImageInsight image and text embeddings grouped on different attributes.}
    \label{fig:pca_groups}
\end{figure}

\begin{figure}[ht]
    \centering
    \begin{subfigure}{.24\columnwidth}
        \centering
         \includegraphics[width=\linewidth]{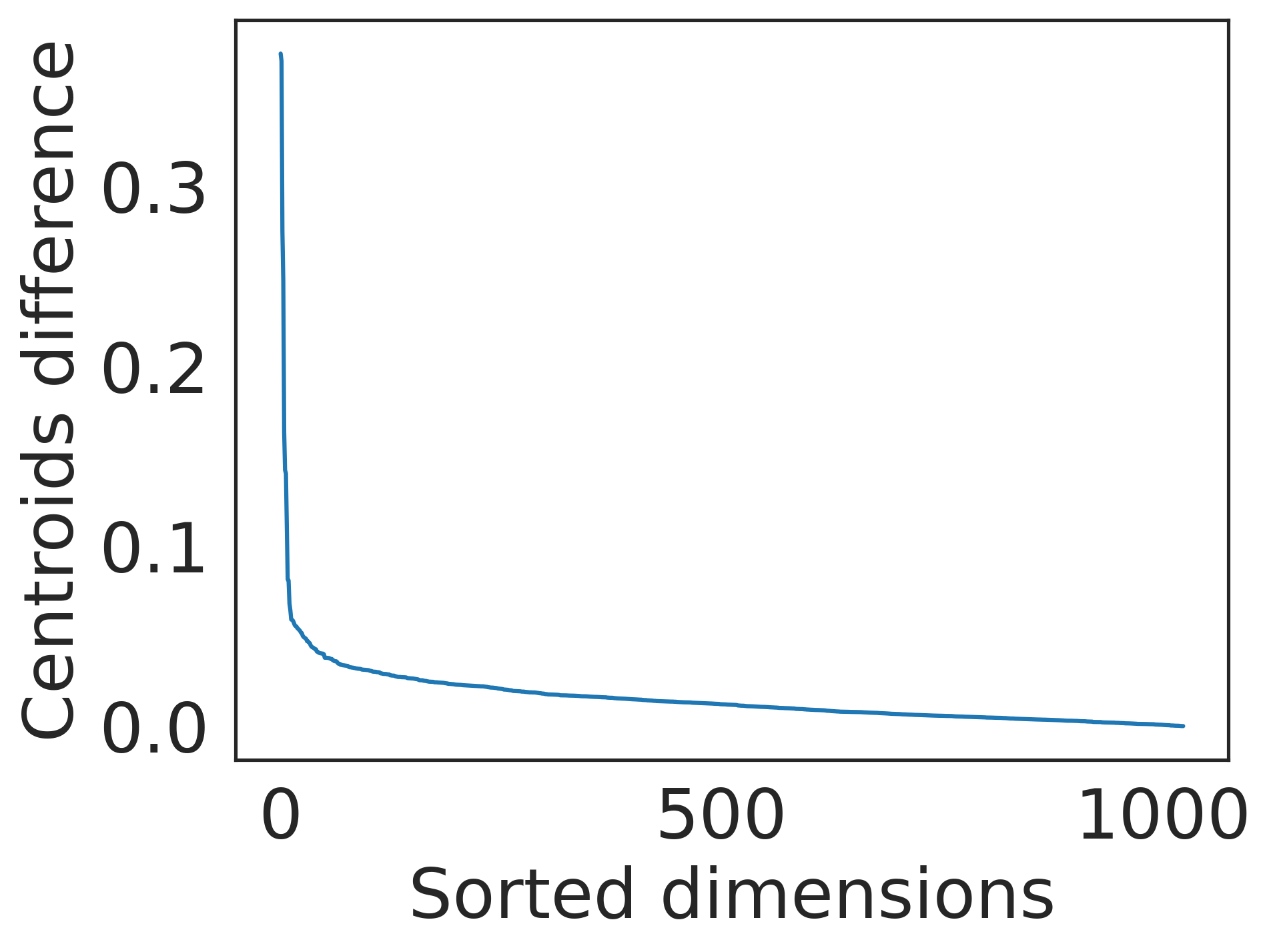}
      \caption{Modality}
      \label{subfig:modality_gap_modality}
    \end{subfigure}
    \begin{subfigure}{.24\columnwidth}
      \centering
      \includegraphics[width=\linewidth]{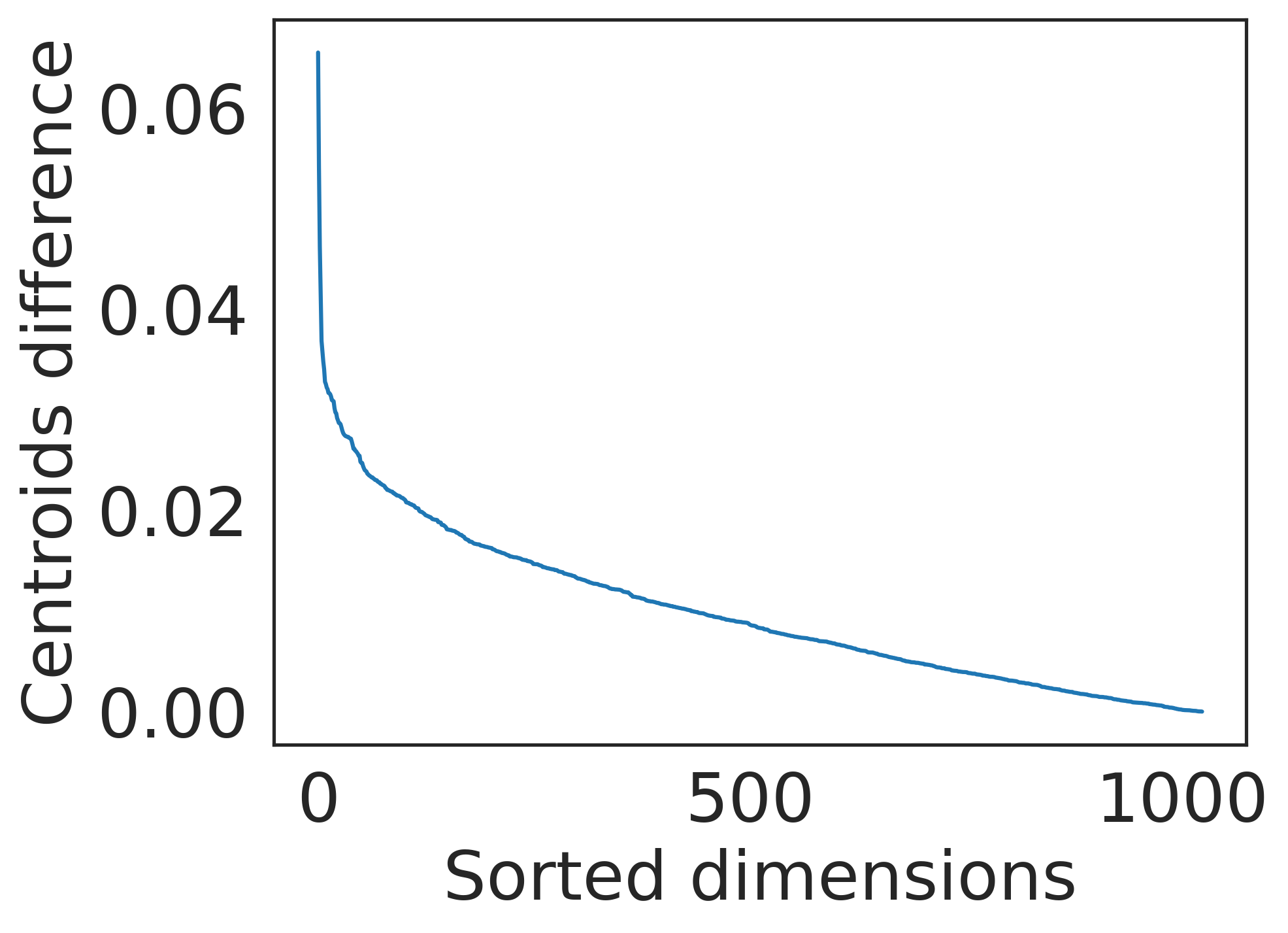}
        \caption{Patient age}
        \label{subfig:modality_gap_age}
    \end{subfigure}
    \begin{subfigure}{.24\columnwidth}
      \centering
      \includegraphics[width=\linewidth]{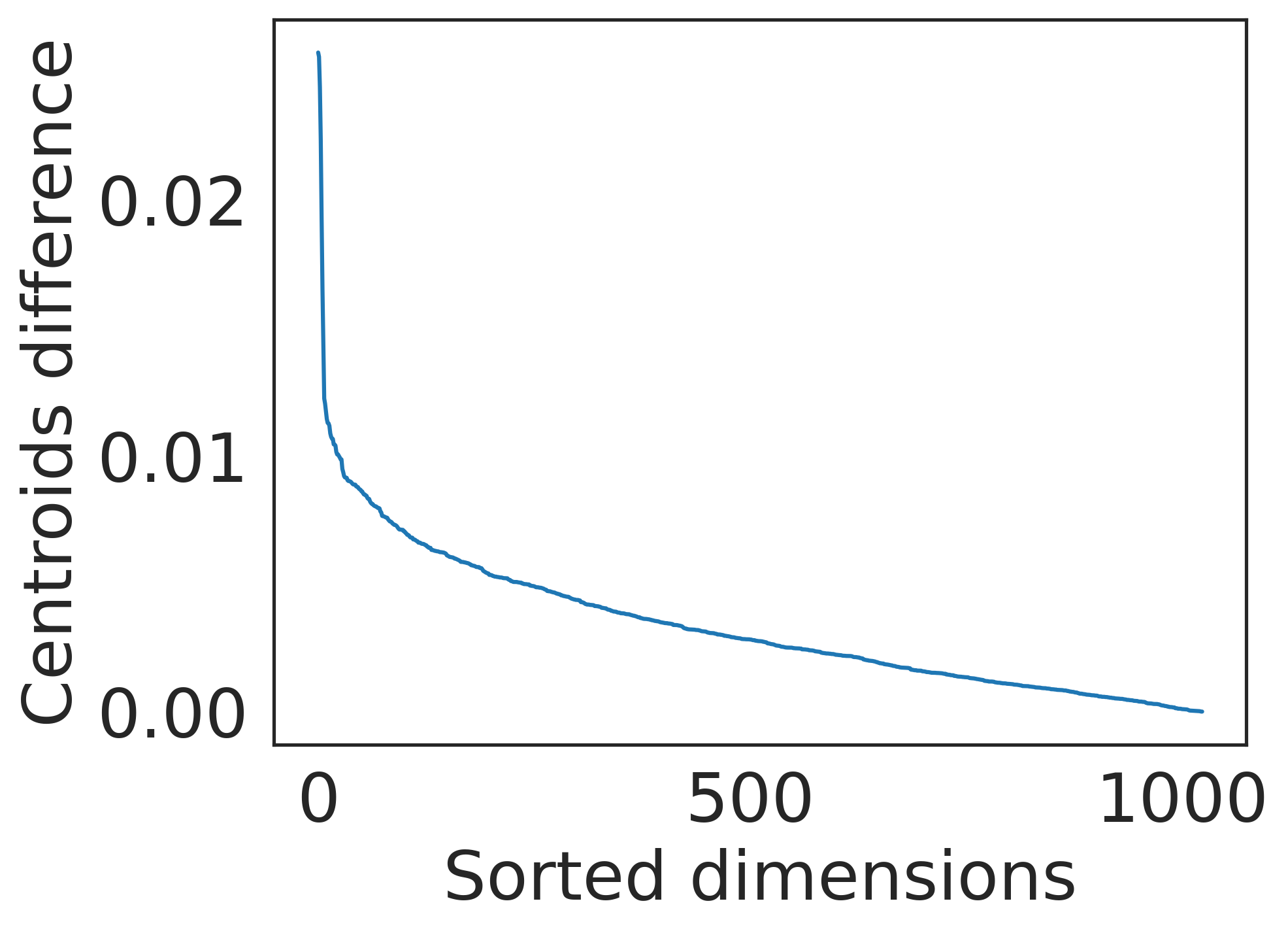}
        \caption{Patient sex}
        \label{subfig:modality_gap_sex}
    \end{subfigure}
    \begin{subfigure}{.24\columnwidth}
      \centering
      \includegraphics[width=\linewidth]{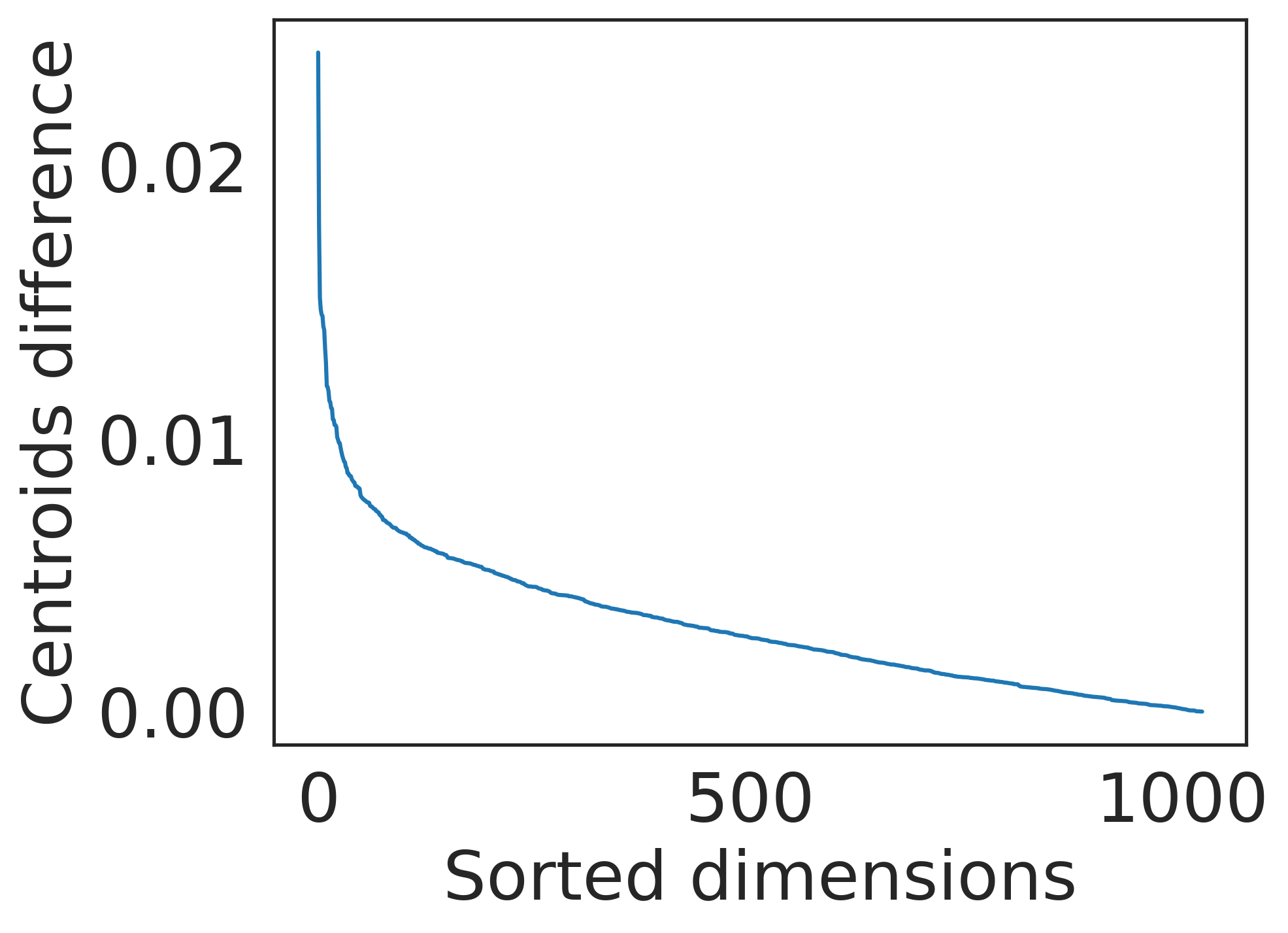}
        \caption{Patient race}
        \label{subfig:modality_gap_race}
    \end{subfigure}
    \caption{Ordered differences between each dimension of the centroids of (a) image and text embeddings generated with MedImageInsight, (b) embeddings of 18-25 years old and 80+ years old patients, (c) female and male patients, and (d) white and black patients. Note that the y-axis range is different in the figures.}
    \label{fig:modality_gap}
\end{figure}

Even though CLIP-based architectures align image and text embeddings using contrastive learning, a simple PCA analysis reveals that in most models (MedImageInsight, CheXzero, CXR-CLIP, and MedCLIP) there is a pronounced gap between the embeddings generated by the image and text encoders. Visible in Fig.~\ref{subfig:modality_gap_pca}, this is aligned with the results from previous studies in natural images~\cite{liang2022mind,schrodi2025two}. Moreover, as shown in~\cite{schrodi2025two}, we also found in Fig.~\ref{subfig:modality_gap_modality} that the gap between the modalities is concentrated on few dimensions.

On the other hand, as shown in Fig.~\ref{subfig:pca_age}-\ref{subfig:pca_race}, we do not see clear patterns in the PCA plots coloured by sensitive attributes. Instead, we observe that the different attributes seem to be well spread across the feature space in both image and text spaces.  
We may conclude from these visualisations that the information is not present in the embedding. However, the differences in subgroup performance observed in the previous section (particularly for age) suggest that certain information related to sensitive attributes may, in fact, be encoded in these representations. To further confirm the algorithmic encoding of protected attributes, we tested the ability of simple supervised models like linear probing, \textit{k}-nearest neighbours, and MLP to classify sensitive attributes from the embeddings for each CLIP-based model and present the results in Table~\ref{tab:sensitive_attribute_classification}. 
While the MLP obtains higher performances, we observe that on patient sex and age all models are able to obtain results above random. 
We can however see that for the patient race, \textit{k}-NN classifiers obtained near-random results for almost all the models and the linear probe is also unable to classify the attribute for some models while the MLP still performs correctly on this attribute. It shows that while it is probably less distinguishable than the other two attributes, it may still be present in the embedding. It is important to note that while such results may show the encoding of information in the embeddings, it is not enough to conclude that they are actually used as shortcuts for other downstream tasks.

As for the modality gap, we analyse the difference between each dimensions of the image embeddings centroids between two subgroup (defined by different sensitive attributes) using only the image modality. Examples are presented in Fig.~\ref{subfig:modality_gap_age}-\ref{subfig:modality_gap_race}. We found that in this case, the differences are much smaller than for the modality gap and more spread across the dimensions. These results suggest that while mitigating the modality gap can be done by focusing on few dimensions, the mitigation of subgroups biases may require more global techniques.

\begin{table}[ht]
	\centering
	\begin{tabular}{c|ccc|ccc|ccc|}
	& \multicolumn{3}{c|}{\textbf{Sex}} & \multicolumn{3}{c|}{\textbf{Race}} & \multicolumn{3}{c|}{\textbf{Age}} \\
	& LP & \textit{k}-NN & MLP & LP & \textit{k}-NN & MLP & LP & \textit{k}-NN & MLP  \\ \hline
	MedCLIP & 0.59 & 0.71 & 0.94 & 0.67 & 0.55 & 0.75 & 0.75 & 0.65 & 0.80  \\
	Biovil & 0.79 & 0.62 & 0.88 & 0.45 & 0.54 & 0.70 & 0.76 & 0.54 & 0.77 \\
	Biovil-t & 0.65 & 0.56 & 0.86 & 0.54 & 0.54 & 0.67 & 0.65 & 0.62 & 0.78  \\
	MedImageInsight & 0.98 & 0.82 & 0.97 & 0.78 & 0.62 & 0.80 & 0.87 & 0.77 & 0.84\\
	CheXzero & 0.75 & 0.65 & 0.93 & 0.66 & 0.53 & 0.69 & 0.76 & 0.69 & 0.78  \\
	CXR-CLIP & 0.97 & 0.89 & 0.97 & 0.82 & 0.55 & 0.80 & 0.84 & 0.58 & 0.80 \\
	\hline
	\end{tabular}
	\caption{Mean AUC of sensitive attributes classification from image embeddings with a linear probe, \textit{k}-NN and MLP}
	\label{tab:sensitive_attribute_classification}
\end{table}

\subsection{Evidence of shortcut learning and miscalibration in CLIP-based models}
\begin{figure}[ht]
    \centering
    \begin{subfigure}{.31\columnwidth}
        \centering
         \includegraphics[width=\linewidth]{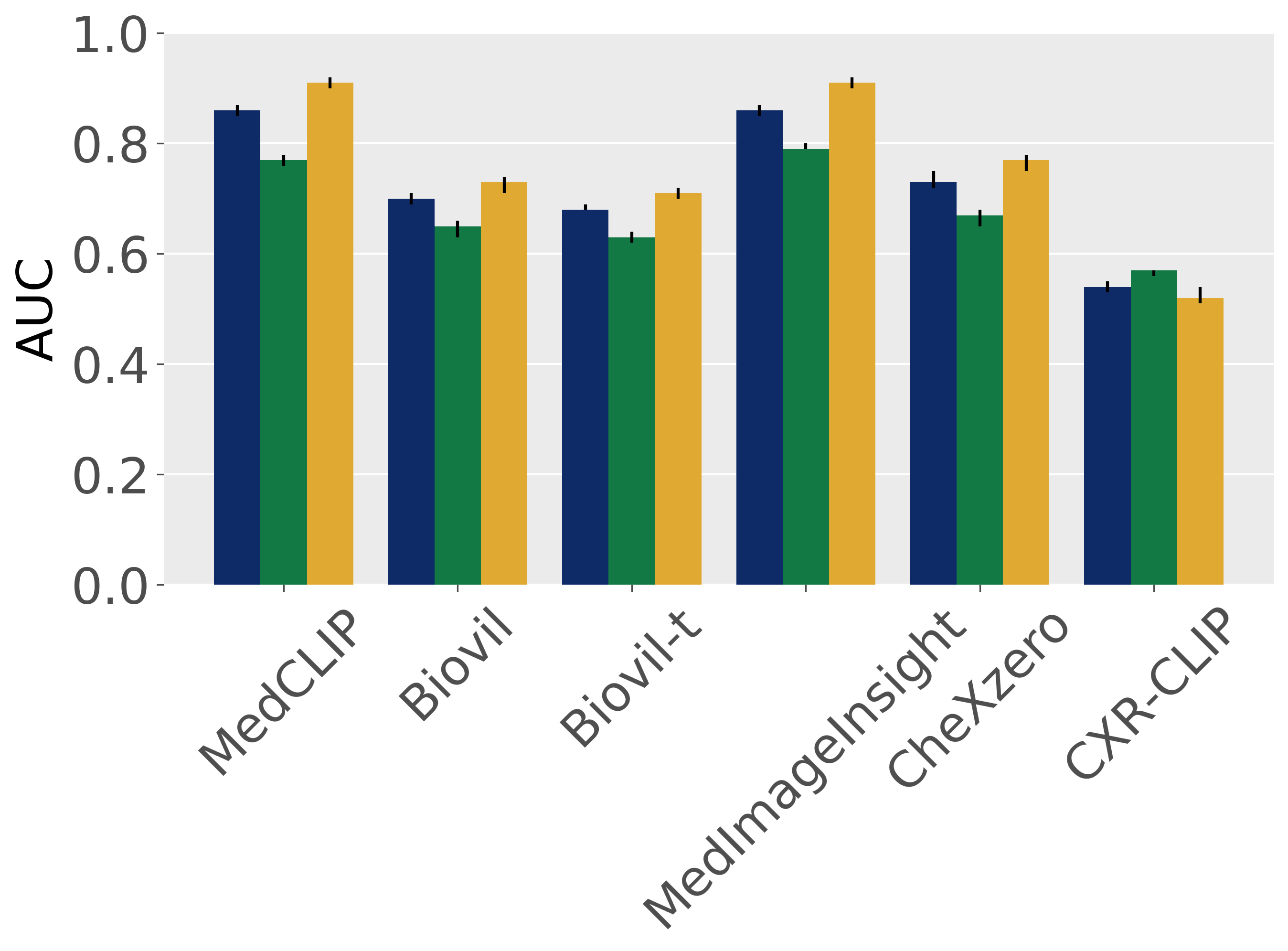}
      \caption{AUC}
      \label{subfig:shortcut_auc}
    \end{subfigure}
    \begin{subfigure}{.31\columnwidth}
      \centering
      \includegraphics[width=\linewidth]{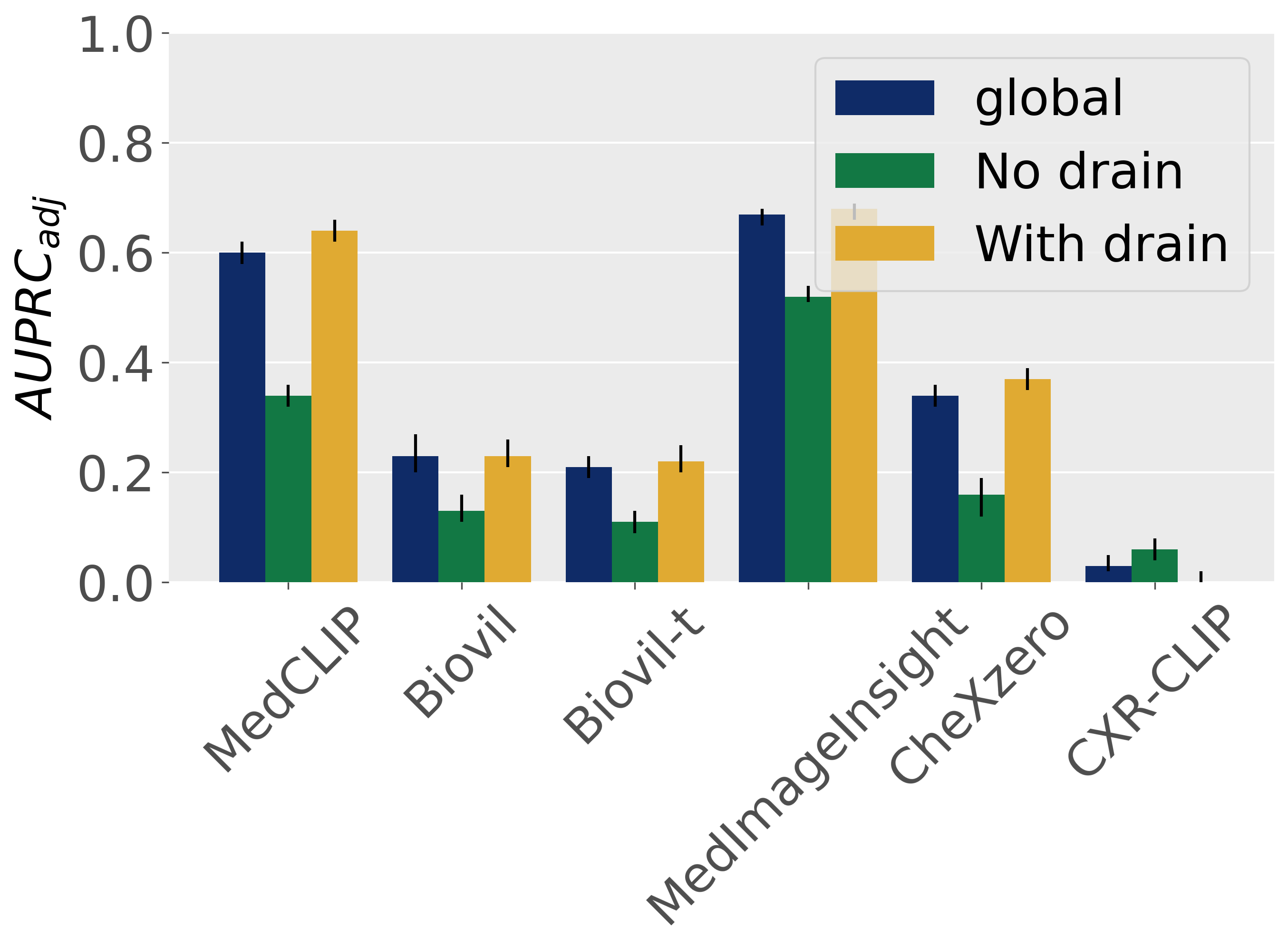}
        \caption{$\text{AUPRC}_{adj}$}
        \label{subfig:shortcut_auprc}
    \end{subfigure}
    \begin{subfigure}{.31\columnwidth}
      \centering
      \includegraphics[width=\linewidth]{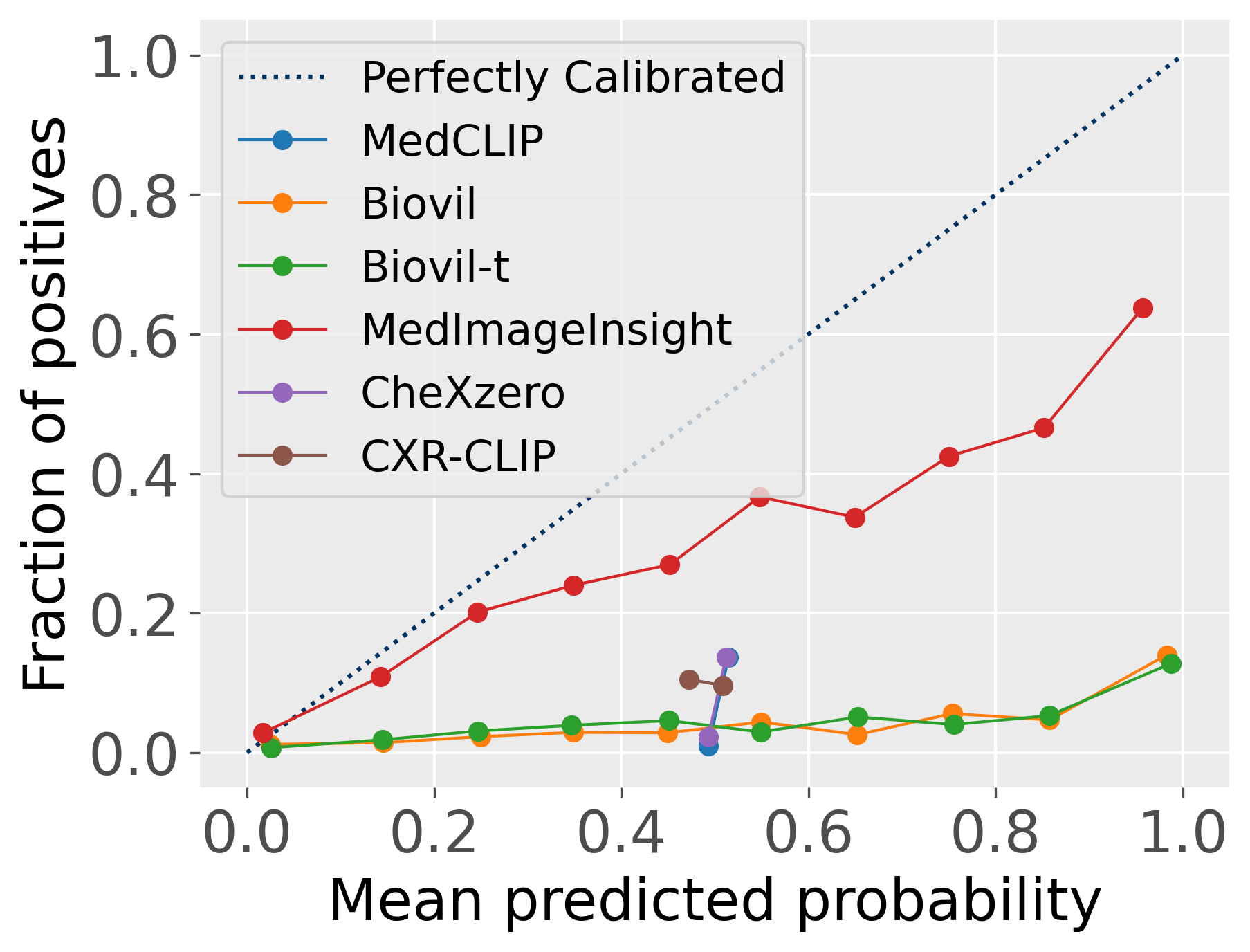}
        \caption{Calibration}
        \label{subfig:calibration}
    \end{subfigure}
    \caption{(a) AUC and (b) adjusted AUPRC of all models on pneumothorax classification of chest X-rays with and without chest drains. (c) Calibration curves of the models on all images.}
    \label{fig:shortcut_results}
\end{figure}

Fig.~\ref{subfig:shortcut_auc} and \ref{subfig:shortcut_auprc} show the results on chest X-ray with and without chest drains. We can see that all models except CXR-CLIP obtain better adjusted AUPRC on images with chest drains compared to X-rays without drains (ranging from +0.09 to +0.30), aligned with previous results on CNN models~\cite{jimenez2023detecting,oakden2020hidden}. Moreover, we present the calibration curves of the models in Fig.~\ref{subfig:calibration}. We see that while MedImageInsight is the most calibrated model, all the models seem to be miscalibrated and overconfident. Interestingly, we can see that for CXR-CLIP, CheXzero and MedCLIP, all the probabilities are around 0.5. Despite this, both MedCLIP and CheXzero achieve acceptable AUC scores, indicating that their discriminative performance remains unaffected, likely because samples are still correctly ranked within this narrow probability range. However, this behaviour significantly complicates the interpretability of individual predictions.

\section{Discussion and conclusions}
In this study, we analysed the fairness of CLIP-based models for chest X-rays across multiple subgroups of patients, showing gaps in the performance obtained on patients of various ages but more balanced results on the other attributes. 
We evaluated the robustness of the models to shortcut learning using chest drains in pneumothorax classification and showed that all models had a better AUPRC$_{adj}$ on images with drains compared to images without drains, indicating their potential reliance on this spurious correlation.
We also assessed the calibrations of the models and found that all models were miscalibrated.
In addition to the performances, we studied the embeddings generated by the models and found that while we could not discover encoding of sensitive attributes using PCA visualizations, such attributes could still be classified from the embeddings using simple supervised models like \textit{k}-NN or MLP, suggesting the encoding of protected attributes.

Fairness and robustness analyses are often constrained by the specific datasets and models used in a given study. Although our experiments span multiple datasets and CLIP-based architectures, the findings may not generalize to all CLIP variants or to other datasets. This highlights the need for similar evaluations in diverse contexts. Furthermore, our fairness analysis focuses on single sensitive attributes at a time, whereas prior research has shown that intersecting subgroups (e.g., age and race) can expose additional fairness concerns~\cite{seyyed2021underdiagnosis}. Lastly, since not all multimodal models follow the CLIP framework, extending such evaluations to generative instead of contrastive multimodal models would provide a more comprehensive understanding.

Our findings, supported by prior work, underscore the importance of improved evaluation frameworks to assess not only overall performance but also fairness across patient subgroups. This requires more diverse datasets, the use of task-appropriate metrics, and going beyond accuracy to consider aspects like calibration and bias.

\subsubsection{Acknowledgements}
\ifdefined\DOUBLEBLIND
***
\else
This work was partially funded by the RHU-EndoVx project (21-RHUS-0011, ANR), Hagnodice project (ANR-21-CE45-0007).
This work was performed using HPC resources from
the Mesocentre computing center of CentraleSupelec. EF was supported by the Google Award for Inclusion Research and a Googler Initiated Grant. EF and MV are supported by the STIC-AmSud CGFLRVE project.
We want to thank the providers of the MIMIC-CXR, NIH-CXR14 and NEATX for creating the datasets used in this study.
\fi
%
%
%
\bibliographystyle{splncs04}
\bibliography{refs}


\end{document}